\pdfoutput=1

\documentclass[11pt,a4paper]{article}
\usepackage[hyperref]{emnlp2020}
\usepackage{times}
\usepackage{latexsym}
\usepackage{amsmath}
\usepackage{graphicx}
\usepackage{multirow}
\usepackage{siunitx}

\usepackage{microtype}

 \aclfinalcopy 

\usepackage{amssymb}

\title{Score Combination for Improved Parallel Corpus Filtering for Low Resource Conditions}

\author{
    Muhammad ElNokrashy $^1$, Amr Hendy$^1$, Mohamed Abdelghaffar$^1$, \\
    {\bf Mohamed Afify$^1$, Ahmed Tawfik$^1$ and Hany Hassan Awadalla$^2$}  \\
    $^1$ Microsoft Egypt Development Center, Cairo, Egypt \\
    $^2$ Microsoft Corporation, Redmond, WA, USA \\
    {\small \tt \{muhammad.elnokrashy,a-amhend,mohamed.abdelghaar,mafify,atawfik,hanyh\}@microsoft.com}
\\
}

\date{}

\begin{document}
\maketitle
\begin{abstract}
This paper describes our submission to the WMT20 sentence filtering task. We combine scores from (1) a custom LASER built for each source language, (2) a classifier built to distinguish positive and negative pairs by semantic alignment, and (3) the original scores included in the task devkit. For the mBART finetuning setup, provided by the organizers, our method shows 7\% and 5\% relative improvement over baseline, in sacreBLEU score on the test set for Pashto and Khmer respectively.
\end{abstract}

\section{Introduction}

Neural machine translation (NMT) brings significant gains to the field of machine translation. However, it is known to be very sensitive to the quality of parallel data \cite{Koehennoisy}. This becomes a serious problem when using large but very noisy corpora for training. There is a lot of work on filtering noisy parallel data. For example, the work in \cite{marcin2018} provides excellent results for large corpora. Low resource languages are even more challenging and the results in \cite{wmt19} using multilingual embeddings are very encouraging.

This paper describes the system submitted to the WMT20 Shared Task on Parallel Corpus Filtering for Low Resource Conditions. Due to time limitation our submission covers only sentence pair filtering. However, some of the proposed techniques could be used for sentence alignment and filtering. The task focuses on the Pashto-English and Khmer-English language pairs. It is required that the participants calculate
scores to sort very noisy parallel sentence pairs provided for each language. The top scoring pairs leading to 5M tokens on the English side are then used to train machine translation for each language pair. The organizers also provide LASER-based scores as a baseline according to the method in \cite{wmt19}, with possibly some  modifications. We describe our general system architecture followed by development experiments to evaluate the merit of different methods and finally report the performance of our best setup per language, using the fairseq recipe provided for the task for both full training (from scratch), and finetuning (mBART) settings. mBART \cite{mBART} is a recently proposed pretraining method. Models initialized using mBART for both language pairs are provided with the task.

Our proposed approach combine the scores provided by the organizers with the following two scores:
\begin{itemize}
    \item Margin distance calculated based on custom LASER built for each language using parallel data and a large amount of forward translated mono data provided by the organizers.    
    \item Cosine distance between the embeddings generated by a classifier taking the pretrained LASER as input and trained to distinguish parallel and non-parallel sentence pairs using parallel data provided in the task.
\end{itemize}
More details on the approach are provided in Section \ref{architecture}. For the mBART setup our method shows 7\% and 5\% relative improvement in sacreBLEU score on the test set for Pashto and Khmer respectively. 

\section{System Architecture}
\label{architecture}
In this section we describe the overall system architecture. We start by presenting language-based preprocessing in Section \ref{preprocess}.The scores provided by the organizers use pretrained LASER \cite{laser} and margin distance \cite{fbmargin1, fbmargin2}. Actually their method obtained state-of-the-art results in WMT19 low resource filtering task for Sinhala, Nepali and Hindi and hence is a very strong baseline. Therefore, we opt to construct sentence embeddings using two different methods, namely custom LASER and classifier-based embeddings, to complement the baseline pretrained LASER. Each embedding method is then used to generate a score for each sentence pair and all the scores are combined to form the final score.
 In the rest of this section, we will describe custom LASER in Section \ref{custom} followed by classifier-based embeddings in Section \ref{classifier} and  finally we outline score combination in Section \ref{combination}.

\subsection{Preprocessing}
\label{preprocess}
We first preprocess the data using language identification on the source side. The results reported in this paper use the BLING \cite{bling} tool from Microsoft but preliminary experiments indicate similar performance using python langid or fasttext \cite{fasttext}. One interesting feature of BLING is that it returns the percentages of different language constituents of a sentence. In the current implementation we keep sentences that have a language identification score of the source language of 80\% or higher. All sentences that do not satisfy the language identification threshold are assigned a very low score and hence are not selected in the final candidate pairs. We will report results without using language identification in the experiments. We generally found it helps a lot for Khmer and actually hurts a bit for Pashto. We expect this is due to the larger Khmer size. Table \ref{langid80percent} shows the original number of sentences and those filtered at 80\% threshold for Pashto and Khmer.

\begin{table}[h]
\centering
 \begin{tabular}{| l | r | r |}
 \hline
 Language & Original & Filtered (kept) \\ [0.5ex] 
 \hline
    Ps     & 1,022,883     &    615,451 (60.17\%) \\
    \hline
    Km     & 4,169,574     &  2,714,664 (65.11\%) \\
\hline
\end{tabular}
\caption{Number of sentences before and after language filtering}
\label{langid80percent}
\end{table}

\subsection{Custom LASER}
\label{custom}
Pretrained LASER \cite{laser} has 93 languages. Some of these languages are under-represented and others, like Pashto, are completely missing. While similar languages tend to help each other it is clearly beneficial to have a custom LASER trained for the languages of interest. In WMT19 results \cite{wmt19}, custom LASER trained on a combination of Hindi, Sinhala and Nepali outperformed the pretrained LASER for the filtering task. Here, we build two separate models for Pashto and Khmer. Since both languages have very different origins we thought it is not beneficial to build a combined model but we haven't verified this experimentally. We use the LASERtrain package \cite{LASERtrain} to train the custom LASER. This package follows the LASER training as given in \cite{laser} and provides experiments on BUCC'18 with good results. It is also possible to fine-tune the pretrained LASER using the languages of interest. We will explore this in future work.
\\\\
Participants in WMT20 are limited to data provided by the organizers. The supplied parallel data for both languages is of rather limited size and is dominated by domain specific data as software localization and religious text and hence we use the provided monolingual text to augment the training data. For each language this is done as follows. We start with the provided sentence scores and filter 5M English tokens as suggested in the task. We use the resulting parallel data to train an mBART initialized MT system.
The sacreBLEU scores on the development test set from the organizers and our internal run are shown in Table \ref{basemBART}.

\begin{table}[h]
\centering
 \begin{tabular}{| l | r | r |} 
 \hline
 Language-pair & Organizers & Internal \\ [0.5ex] 
 \hline
    Ps-En & 12.2 & 11.6 \\
    \hline
    Km-En & 10.6 & 10.4 \\
\hline
\end{tabular}
\caption{\label{basemBART} SacreBLEU for mBART on development test set as provided by the organizers and for internal run.}
\end{table}

In addition to the noisy parallel sentences, the organizers provide additional parallel data and monolingual data for both languages. The parallel data comes mainly from OPUS and consists of 290K and 123K pairs for Khmer and Pashto respectively. The monolingual data for Khmer has around 13M sentences while that of Pashto has around 6M sentences. For more information about the sources of these data we refer the reader to \cite{wmt2020}.
\\
The resulting internal system is used to forward translate various amounts of monolingual data from the source language into English. We found in preliminary experiments that using around 3M monolingual sentences gives good performance (more on the evaluation below). These sentences are randomly selected from the monoingual data. The synthetic pairs for Pashto-English and Khmer-English are used to augment the provided clean parallel data to train the custom LASER for each language. The reason we build unidirectional custom LASER is that most of our data is synthetic and the performance of the translation in the opposite direction English-Pashto(Khmer) is expected to be quite poor. In addition, English is very well represented in the pretrained LASER.
\\
It is good to have a way to evaluate the quality of the built custom LASER. Building machine translation (MT) every time is very costly. To this end, we use a BUCC-like setup to evaluate the quality of the custom embeddings. We use the development set for each language pair. For each sentence on the source side we use the corresponding embedding to find the nearest neighbor, based on cosine distance, on the target side and calculate the top-1 accuracy. We do the same in the other direction (target to source) and average both numbers.
\begin{equation}
\begin{split}
    \operatorname{BUCC}(S)
    = \frac{1}{2 \lvert S \rvert} \, \Biggl(
            & \sum_i{\mathbb{I}\left(\arg\max_j{S_{i,j}} = i\right)} \\
            + & \sum_j{\mathbb{I}\left(\arg\max_i{S_{i,j}} = j\right)}
        \Biggr)
\end{split}
\end{equation}
where $S$ is the matrix of pairwise similarity scores for pairs in the source-target development set.\\
Accuracies using pretrained and custom LASER for Pashto and Khmer are shown in Table \ref{accuracylaser}.

\begin{table}[h]
\centering
 \begin{tabular}{| l | r | r |} 
 \hline
 Language-pair & Pretrained & Custom \\ [0.5ex] 
 \hline
    Ps-En & 9.56\% & 31.97\% \\
    \hline
    Km-En & 1.04\% & 39.50\% \\
\hline
\end{tabular}
\caption{\label{accuracylaser} BUCC-like accuracy scores on devtest set of the filtration task}
\end{table}
\vspace{-0.1em}

Once the custom LASER of a language is trained it is used to calculate the score of a sentence pair using the margin distance as shown in Equation \ref{margineqn}. The margin is implemented efficiently using \cite{fbmargin2}.

\begin{equation}
\begin{split}
\operatorname{score}(x, y)
    = & \operatorname{margin}\Bigl(
        \cos(x, y), \\
        0.5 \, \bigl(
            &\operatorname*{mean}\left\{ \cos(z, x) \mid z \in \operatorname{NN}_k(x) \right\} \\
            + &\operatorname*{mean}\left\{ \cos(z, y) \mid z \in \operatorname{NN}_k(y) \right\}
            \bigr)
    \Bigr)
\end{split}
\label{margineqn}
\end{equation}

\subsection{Classifier-Based Scores}
\label{classifier}
In addition to custom LASER presented in the previous section we use scores provided from a classifier
trained to distinguish parallel and non-parallel sentence pairs. It takes pretrained LASER embeddings of a sentence pair $u$ and $v$ and transforms them using a fully-connected layer with ReLU non-linearity. Similar to \cite{sentemb} it inputs the concatenation 
$[u_{tr}; v_{tr}; \left|u_{tr}-v_{tr}\right|]$, where $u_{tr}$ and $v_{tr}$ are the outputs of the fully connected layer, to a softmax classifier with two outputs representing the positive and negative pairs. The network is trained using the cross-entropy criterion. During testing, LASER embeddings of a sentence
pair are passed through the fully connected layer and their cosine distance is calculated as the required score. The rationale is that the transformed embeddings provide better representation to separate positive and negative pairs compared to pretrained LASER.

For each language, the classifier is trained on the positive pairs provided by the organizers. Following \cite{sentencefilteringacl} for each sentence the negative pair is selected at random from the following:
\begin{itemize}
\item Select a sentence from its adjacent sentences within a window size of k 
(where k = 2 in our experiments).
\item Truncate 30-70\% words of the sentence.
\item Swap the order of 30-70\% of the words of the sentence.
\end{itemize}
After forming the positive and negative data around 500 example pairs, per language, are kept as validation set. The classification accuracy, on the validation set,  for Pashto is 97\% while that of Khmer is 98.5\%. 

\subsection{Score Combination}
\label{combination}
Based on the previous sections each input sentence pair $x, y$ has three scores. Assume the pretrained
LASER embeddings are $x_{p}$ and $y_{p}$ and the custom LASER embeddings are  $x_{c}$ and $y_{c}$. We can write the combined score $S(x, y)$ as follows:
\begin{equation}
 S(x, y) = S_{mg}(x_{p},y_{p}) + S_{mg}(x_{c},y_{c}) + S_{cl}(x_{p},y_{p})
 \label{eqncombination}
\end{equation}  
where $S_{mg}()$ indicates margin distance and $S_{cl}()$ indicates classifier distance.
We choose to use a simple sum instead of using trainable weights because the provided parallel data that could be used to train the weights is very specific and could result in biased estimates of the weights.

We also experiment with minimum-maximum normalization that we found very useful in the case of Khmer. For this normalization each component score in Equation \ref{eqncombination} is modified as follows:
\begin{equation}
S_{norm} = \frac{S-S_{min}}{S_{max}-S_{min}}
\label{mmequation}
\end{equation}
where $S_{min}$ and $S_{max}$ are the minimum and maximum scores over all the pairs.

\section{Experimental Results}
In this section we first present the results of various experiments to arrive at the final system architecture for both languages in Section \ref{devexperiments}. An internal system with a small architecture is used for fast turn-around. This is followed by running experiments with the final architecture using the official scripts provided by the organizers for both the from scratch and mBART settings. The training parameters are described in Table \ref{tab:mbart-train}. Training was done using the fairseq framework \cite{ott2019fairseq}.

\begin{table}[h]
\centering
\begin{tabular}{|l|l|}
\hline
\textbf{Parameter} & \textbf{Value} \\ \hline
\textit{Optimizer}    & ADAM \cite{adam}                         \\ \hline
\textit{LR}           & $\num{1e-4}$                             \\ \hline
\textit{LR Scheduler} & Inverse Sqrt                             \\ \hline
\textit{LR Warmup}    & $\num{1e-7}$                             \\ \hline
\textit{Warmup Steps} & 4000 updates, linear                     \\ \hline
\textit{Batch Size}   & 16K tokens                               \\ \hline
\textit{Early Stop}   & 10 stalls of 5000 updates each           \\ \hline
\end{tabular}%
\caption{Finetuning parameters for mBART}
\label{tab:mbart-train}
\end{table}

\subsection{Development Experiments}
\label{devexperiments}
This section outlines various development experiments for Pashto and Khmer. As mentioned above an internal system with a small configuration is used to compare different configurations. The sacreBLEU
scores for Pashto are shown in Table \ref{pashtodevtab} while those for Khmer are in Table \ref{khmerdevtab}. 

\begin{table}[h]
\centering
 \begin{tabular}{| l | r | r |} 
 \hline
                & Dev. Set  & Test Set     \\ \hline
B               & 7.4       & 8.4          \\ \hline
C               & 9.3       & 9.3          \\ \hline
B + C           & 9.2       & 10.4         \\ \hline
B + C + Cl      & 9.5       & 10.5         \\ \hline
B + C + Cl (BL) & 8.3       & 9.6          \\ \hline
\end{tabular}
\caption{\label{pashtodevtab} Pashto Development results (in SacreBLEU) for different configurations on development and test sets. B stands for baseline, C for custom, Cl for classifier and BL for BLING}
\end{table}

\begin{table}[h]
\centering
 \begin{tabular}{| l | r | r |} 
 \hline
                  & Dev. Set & Test Set     \\ \hline
B          & 8.8      & 7.0   \\ \hline
C            & 4.3      & 3.6   \\ \hline
C (BL)            & 6.5      & 5.2   \\ \hline
B + C (BL)        & 9.5      & 7.6   \\ \hline
B + C + Cl (BL)   & 9.9    &  8     \\ \hline
\end{tabular}
\caption{\label{khmerdevtab} Khmer Development results (in SacreBLEU) for different configurations on development and test sets.B stands for baseline, C for custom, Cl for classifier and BL for BLING.B+C+Cl result for Khmer uses minimum-maximum normalization.}
\end{table}

From the two tables we can see that there is some significant difference in behavior between Pashto and Khmer. This can be summarized as follows:
\begin{itemize}
    \item While custom LASER is better than pretrained LASER for Pashto it is worse for Khmer. We attribute this to the existence of Khmer and absence of Pashto in pretrained LASER. For Pashto, even with some small parallel data and synthetic data we can see some nice gains.
    \item BLING language filtering is crucial for Khmer while it hurts a bit for Pashto. We attribute this to the larger size and the noisier nature (from the view point of having more English words in the source side) of the Khmer data.
    \item Even if the custom LASER for Khmer is significantly worse than the pretrained one it helps when combined with the baseline.
\end{itemize}

\subsection{Final Experiments}
Based on the above observations, we decided to have our configurations for the final experiments of the two languages as follows: 
\begin{itemize}
    \item Use BLING filtering for Khmer but not for Pashto. 
    \item Use the combined scores of pretrained LASER, custom LASER and classifier for both Pashto and Khmer.
    \item Experiment with and without min-max normalization.
\end{itemize}

The results using both from scratch (Full) and mBART (FT) settings as supplied by the organizers are shown in Table \ref{tbl:results}.

\begin{table}[ht]
\centering
\begin{tabular}{|l|l|r|r|r|}
\hline
\multirow{2}{*}{}      & \multirow{2}{*}{Mode}     & \multicolumn{1}{c|}{\multirow{2}{*}{Base}}     & \multicolumn{2}{c|}{Ensemble}                               \\ \cline{4-5} 
                       &                           & \multicolumn{1}{l|}{}                          & \multicolumn{1}{l|}{min-max} & \multicolumn{1}{l|}{No norm} \\ \hline
\multirow{2}{*}{Ps-En} & Full                      &    10.04                                       &   10.12                      &    10.05                     \\ \cline{2-5} 
                       & FT                        &    11.61                                       &   11.99                      &    12.38                     \\ \hline
\multirow{2}{*}{Km-En} & Full                      &    7.16                                        &   7.88                       &    6.36                      \\ \cline{2-5} 
                       & FT                        &    10.56                                       &   11.12                      &    9.02                      \\ \hline
\end{tabular}
\caption{\label{tbl:results} Final results in SacreBLEU on devtest set. Full stands for from Scratch and FT for mBART.}
\end{table}

Based on the results in the table our submission used minimum-maximum normalization for Khmer but not for Pashto. By looking into the unnormalized scores we found that for Khmer they tend to be dominated by the classifier score, undermining both the baseline and custom LASER, and hence the normalization helps to
bring all scores to the same dynamic range.

\section{Summary}
This paper present the description of our submission to WMT20 sentence filtering task. By building custom LASER and a classifier to distinguish positive and negative pairs. For the mBART setup our method shows 7\% and 5\% relative improvement in sacreBLEU score on the test set for Pashto and Khmer respectively.There are a lot of extensions along the proposed directions to improve sentence filtering for the low resource setting.

\bibliography{paper}
\bibliographystyle{acl_natbib}

\end{document}